\documentclass[conference]{IEEEtran}
\IEEEoverridecommandlockouts
\usepackage{cite}
\usepackage{amsmath,amssymb,amsfonts}
\usepackage{algorithmic}
\usepackage[ruled,vlined]{algorithm2e}
\usepackage{graphicx}
\usepackage{textcomp}
\usepackage{xcolor}
\usepackage{lipsum}
\usepackage{graphicx}
\usepackage{comment}
\usepackage{amsmath,amssymb} 
\usepackage{color}
\usepackage{cite}
\usepackage{multirow}
\usepackage{wrapfig,lipsum,booktabs}
\usepackage{makecell}
\usepackage{tabularx}
\usepackage{float}
\usepackage{stfloats}
\usepackage{tabu}
\usepackage[a4paper, total={184mm,239mm}]{geometry}
\mathchardef\simsym"0218
\def\BibTeX{{\rm B\kern-.05em{\sc i\kern-.025em b}\kern-.08em
    T\kern-.1667em\lower.7ex\hbox{E}\kern-.125emX}}
    
\begin{document}

\title{Efficiency-driven Hardware Optimization for Adversarially Robust Neural Networks}

\author{Abhiroop Bhattacharjee$^{*}$\thanks{\hspace{-3mm}$^*$ These authors have contributed equally to this work.}, Abhishek Moitra$^{*}$, and Priyadarshini Panda \small
\\ \{abhiroop.bhattacharjee, abhishek.moitra, priya.panda\}@yale.edu \\ Department of Electrical Engineering, Yale University, USA \normalsize
}

\maketitle

\begin{abstract}

With a growing need to enable intelligence in embedded devices in the Internet of Things (IoT) era, secure hardware implementation of Deep Neural Networks (DNNs) has become imperative. We will focus on how to address adversarial robustness for DNNs through efficiency-driven hardware optimizations. Since memory (specifically, dot-product operations) is a key energy-spending component for DNNs, hardware approaches in the past have focused on optimizing the memory. One such approach is approximate digital CMOS memories with hybrid 6T-8T SRAM cells that enable supply voltage (Vdd) scaling yielding low-power operation, without significantly affecting the performance due to read/write failures incurred in the 6T cells. In this paper, we show how the bit-errors in the 6T cells of hybrid 6T-8T memories minimize the adversarial perturbations in a DNN. Essentially, we find that for different configurations of 8T-6T ratios and scaled Vdd operation, noise incurred in the hybrid memory architectures is bound within specific limits. This hardware noise can potentially interfere in the creation of adversarial attacks in DNNs yielding robustness. Another memory optimization approach involves using analog memristive crossbars that perform Matrix-Vector-Multiplications (MVMs) efficiently with low energy and area requirements. However, crossbars generally suffer from intrinsic non-idealities that cause errors in performing MVMs, leading to degradation in the accuracy of the DNNs. We will show how the intrinsic hardware variations manifested through crossbar non-idealities yield adversarial robustness to the mapped DNNs without any additional optimization.

\end{abstract}


\section{Introduction}
Today, Deep Neural Networks (DNNs) are becoming ubiquitous tools for autonomous computation and edge classification tasks. However, adversarial attacks have been shown to severely degrade the performance of DNNs by adding minute, undetectable perturbations into the input images. This prevents their deployment in critical environments like aircraft and medical diagnostics. 

Several algorithmic defenses have been proposed for resisting adversarial attacks. These involve input randomization, image compression, and Gaussian noise augmentation \cite{compression, resize, he2019parametric}. Of all the works, adversarial training has been shown to be the best defense against adversarial attacks. Here, the network is trained with clean and adversarial images chosen precisely to offer a defense against all adversarial attacks\cite{pgd}. The authors in \cite{MIT_survey} describe popular methods that have been shown to improve the adversarial robustness of DNNs.

Apart from algorithmic defenses, there is a recent body of efficiency-centric techniques~\cite{pixeld, defquant, quanos} to improve robustness of neural networks against adversarial perturbations. The authors in~\cite{pixeld} perform discretization of the input space (i.e. restrict allowed pixel levels from 256 values or 8-bit to 4-bit, 2-bit) to improve the adversarial robustness of DNNs for a substantial range of perturbations, besides improvement in its computational efficiency with minimal loss in test accuracy. Likewise, \textit{QUANOS}~\cite{quanos} is a framework that performs layer-specific hybrid quantization of DNNs based on a metric termed as \textit{Adversarial Noise Sensitivity} (ANS) to generate energy-efficient, accurate and adversarially robust models. Thus, previous works~\cite{pixeld, quanos, defquant} show that efficiency-driven hardware optimization techniques can be leveraged to improve software vulnerability, such as, adversarial attacks, while yielding energy-efficiency. However, in this work we find that the intrinsic hardware noise or non-idealities can inadvertently improve adversarial security of DNNs without any additional optimization. 

In the light of efficiency-driven optimization, approximate computing leverages the error tolerance capability of DNNs offering energy efficient computation with slight or no decrease in performance. Similarly, approximate memories like hybrid 8T-6T SRAM memories require low read and write access energy making the design energy efficient \cite{bortolotti2014approximate, chang2011priority}. However, at very low voltages, they are prone to high bit-error rates that lead to a large dip in the DNN performance \cite{srinivasan2016significance}. In this work, we show that these infamous bit-errors can be controlled and introduced into strategically selected DNN layers to improve the adversarial robustness of the DNN. This is done by configuring the hybrid memory configurations: 8T-6T cell ratios and supply voltage \cite{moitra2020exposing}. 

In the recent years, memristive crossbars are being increasingly used to implement DNNs by effciently computing analog dot-products~\cite{schuman, mo-rram, sharad}. The crossbar synapses have been widely realized using various emerging technologies such as, \textit{Resistive RAM} (ReRAM), \textit{Phase Change Memory} (PCM) and Spintronic devices~\cite{chen, sengupta, sttsnn}. These devices exhibit high on-chip storage density, non-volatility, low leakage and low-voltage operation and thus, enable compact and energy-efficient implementation of DNNs~\cite{puma, rxnn}. Nevertheless, the analog nature of computing dot-products in crossbars has certain limitations owing to device-level and circuit-level non-idealities such as, interconnect parasitics, process variations in the synaptic devices, driver and sensing resistances, etc.~\cite{geniex,rxnn,cxdnn}. These non-idealities manifest as errors in the analog dot-product computations in the crossbars, thereby adversely affecting the computational accuracy of DNNs~\cite{lee2020learning,cxdnn}. Numerous frameworks have been developed in the past to model the impact of such non-idealities and accordingly, retraining the synapses to mitigate accuracy losses~\cite{cxdnn, xchangr, indranil, lee2020learning, geniex}. However, an interesting aspect of these non-idealities in providing resilience to DNNs against adversarial attacks has received little attention. In this work, we bring out the fact that a DNN model mapped on crossbars, while suffering accuracy degradation, can be more adversarially robust than the software baseline~\cite{bhatta}. 

In summary, the key contributions of this work are as follows:
\begin{itemize}
\item We show that bit-error noise due to 6T-SRAM cells in hybrid memories can be leveraged to improve the adversarial robustness of the DNN model. Additionally, we show that the bit-error noise depends on the hybrid memory configurations (8T-6T ratios and supply voltage). Using this, we propose a methodology to select the most suitable DNN layers for noise injection and the corresponding amounts of bit-error noise. 
\item To validate our findings, we perform experiments using CIFAR10 and CIFAR100 datasets on both VGG19 and ResNet18 networks. We launch FGSM attacks of various strengths in order to test the effectiveness of our methodology.
\item We study the benefits conferred upon by crossbar non-idealities in terms of adversarial robustness in DNNs by employing a systematic framework in \textit{PyTorch} to map DNNs onto memristive crossbars. We conduct experiments on state-of-the-art neural networks (VGG8 \& VGG16) using benchmark datasets (CIFAR-10 \& CIFAR-100) and show that non-idealities lead to higher adversarial robustness ($\sim10-20\%$ for FGSM and PGD attacks) for DNNs on crossbars than baseline software models.
\item We further present a comparison of non-ideality driven adversarial defense with other state-of-the-art efficiency-driven quantization techniques, \textit{viz.} 4-bit discretization of input pixels~\cite{pixeld} and \textit{QUANOS}~\cite{quanos} to illustrate the efficacy of the approach in addressing adversarial vulnerabilities. 

\end{itemize}

\section{Background and Motivation}

\subsection{Adversarial Attacks}
\label{sec:adv}

Adversarial attacks are those in which a DNN gets fooled by applying structured but small perturbations to the inputs, leading to high confidence misclassification~\cite{quanos}. The authors in~\cite{goodfellow} have proposed a method called \textit{Fast Gradient Sign Method} (FGSM) to generate the adversarial input by linearization of the loss function ($L$) of the trained models with respect to the input ($X$) as shown in equation~(\ref{eq:Xadv}).
\begin{equation} \label{eq:Xadv}
X_{adv}~=~X~+~\epsilon~\times~sign(\nabla_{x} L(\theta,X,y_{true}))
\end{equation}
Here, $y_{true}$ is the true class label for the input X; $\theta$ denotes
the model parameters (weights, biases etc.) and $\epsilon$ quantifies the strength of the perturbation added. 

The quantity $\Delta=\epsilon~\times~sign(\nabla_{x} L(\theta,X,y_{true}))$ is the net perturbation added to the input ($X$). It is noteworthy that gradient propagation is, thus, a crucial step in unleashing an adversarial attack. Furthermore, the contribution of gradient to $\Delta$ would vary for different layers of the network depending upon the activations~\cite{quanos}. In addition to FGSM attacks, multi-step variants of FGSM, such as \textit{Projected Gradient Descent} (PGD)~\cite{pgd} have also been proposed that cast stronger attacks. 

\begin{figure}[t]
    \centering
    \includegraphics[width=0.8\linewidth]{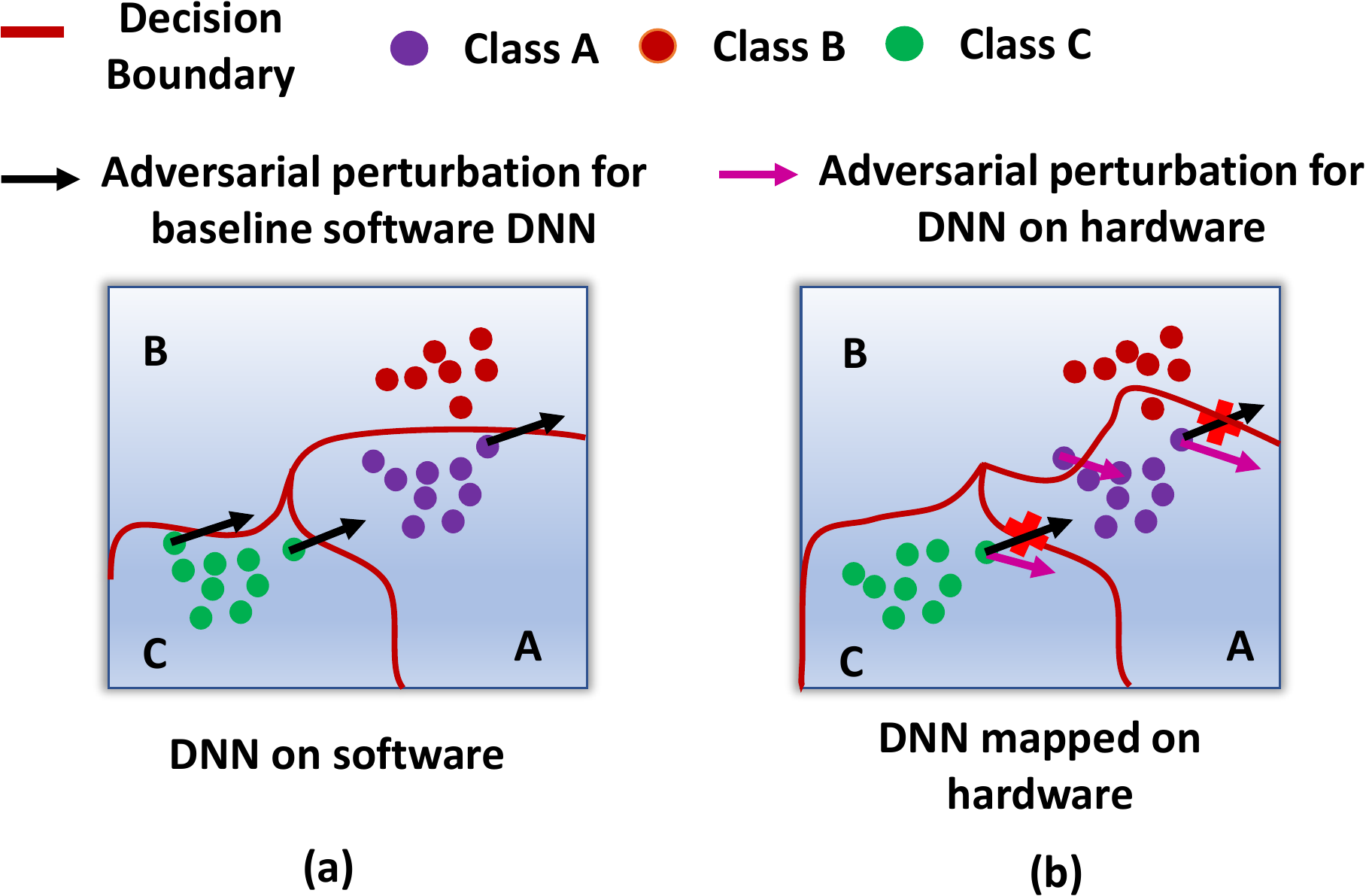}%
    \caption{Pictorial depiction of creation of adversaries for software and hardware-based DNNs - (a) The data points (shown as `dots’) encompass
the data manifold in the high-dimensional subspace. The classifier is trained to separate the data into different classes or hypervolumes based on which the decision boundary is formed. Adversaries are created by perturbing the data points into these empty regions or hypervolumes and are thus misclassified; (b) The decision boundaries get shifted owing to the intrinsic noise in hardware, resulting in the placement of certain data points into a different hypervolume leading to accuracy degradation. However, many data points remain restricted in their original hypervolumes on adversarial perturbations. This results in better adversarial robustness in hardware-based DNNs.}
    \label{picadv}
    \vspace{-4mm}
\end{figure}

To build resilience against small adversarial perturbations, defense mechanisms such as gradient masking or obfuscation~\cite{gradmask} have been proposed. Such methods construct a model devoid of useful gradients, thereby making it difficult to create an adversarial attack. In this work, we show that when DNNs are mapped onto hardware, the intrinsic hardware noise (or non-idealities) inadvertently lead to defense via gradient obfuscation against adversarial perturbations. The entire phenomenon has been summarized in \figurename{~\ref{picadv}}. 

In this work, \textit{Clean Accuracy} refers to the accuracy of a DNN when presented with the test dataset in absence of an adversarial attack. We define \textit{Adversarial Accuracy} as the accuracy of a DNN on the adversarial dataset created using the test data for a given task. \textit{Adversarial Loss} ($AL$) is defined as the difference between clean and adversarial accuracies. Smaller the value of AL, greater the robustness of the DNN.

\subsection{Bit-error noise in hybrid 8T-6T memories}
In 6T-SRAM cells, with voltage scaling, the time to write into the SRAM cell and the time to read from the SRAM cell increases. This leads to an increase in the bit-error rate at scaled voltages \cite{mukhopadhyay2005modeling}. These bit errors lead to a bit-error noise in hybrid 8T-6T memory architectures which we leverage to improve the adversarial robustness of DNNs. To estimate the amount of bit-error noise, we design a 6T-SRAM cell using 22nm predictive model, having static read noise and write noise margins of 195mV and 250mV, respectively.
\begin{figure}[t]
  \centering
    \includegraphics[width=0.3\textwidth]{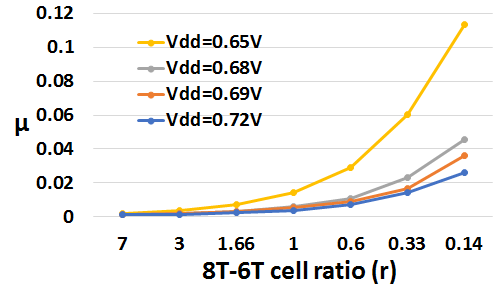}
  
  \caption{Variation of average surgical noise perturbation $\mu$ with 8T-6T SRAM cell ratio $r$ for different supply voltages $V_{DD}$}
  \label{sn_vs_r_vs_vdd}
\end{figure}

Next, using the procedure employed in \cite{srinivasan2016significance}, we calculate bit-error values corresponding to different scaled voltages. These bit-error values are used to model the amount of bit-error noise in hybrid 8T-6T memories at different 8T-6T ratios $r$ ($\frac{\#8T~cells}{\#6T~cells}$) and supply voltages $V_{DD}$. Our experiments shown in Fig. \ref{sn_vs_r_vs_vdd}, corroborate that the bit-error noise can be estimated using the value of $r$ and $V_{DD}$. The amount of bit-error noise ($\mu$), increases as the number of 6T cells increase. Moreover, the overall bit-error noise increases with higher voltage scaling.

\subsection{Non-idealities in analog crossbar arrays}

\begin{figure*}[t]
    \centering
    \includegraphics[width=0.9\linewidth]{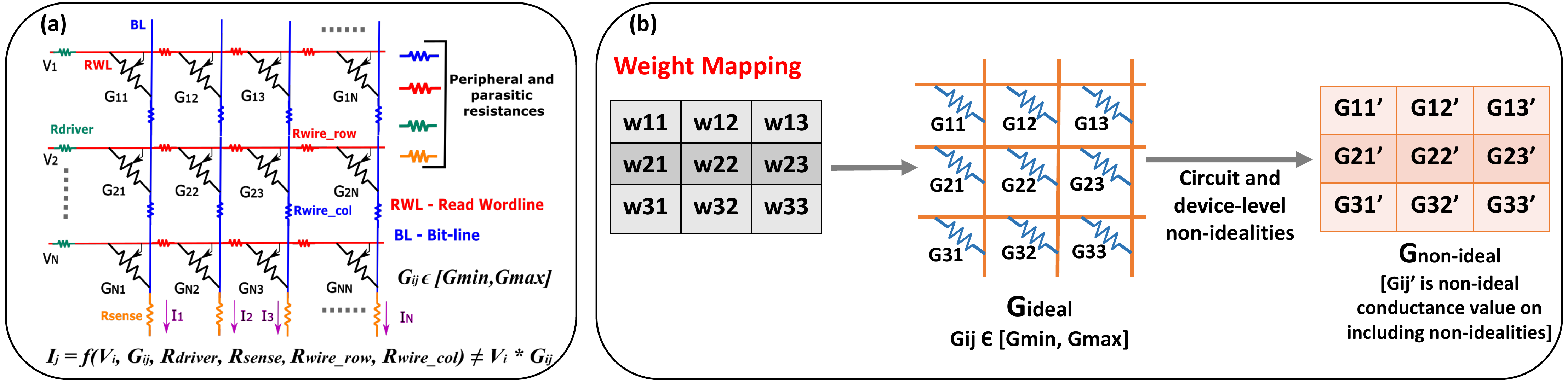}%
    \caption{(a) An NxN crossbar array with the various resistive non-idealities marked. The non-idealities lead to imprecise dot-product currents and that manifests as accuracy degradation when DNNs are evaluated on crossbars; (b) A diagrammatic representation of the mapping of a 3x3 weight matrix onto a 3x3 crossbar array having resistive non-idealities and process variation in the synaptic devices}
    \label{xbar}
\end{figure*}

DNNs can be implemented on memristive crossbars wherein, the activations are mapped as analog voltages $V_i$ input to each row and weights are programmed as synaptic device conductances ($G_{ij}$) at the cross-points as shown in \figurename{~\ref{xbar}} (a). For an ideal crossbar array, during inference, the voltages interact with the device conductances and produce a current (governed by Ohm's Law). Consequently, by Kirchhoff's current law, the net output current sensed at each column $j$ is the sum of currents through each device, \textit{i.e.} $I_{j(ideal)} = \Sigma_{i=1}^{N}{G_{ij} * V_i}$. We term the matrix $G_{ideal}$ as the collection of all $G_{ij}$s for a crossbar instance. However, in reality, the analog nature of the computation leads to various non-idealities, such as, circuit-level resistive non-idealities in the crossbars and device-level variations. 

\figurename{~\ref{xbar}}(a) describes the equivalent circuit for a crossbar accounting for non-idealities, \textit{viz.} $Rdriver$, $Rwire\_row$, $Rwire\_col$ and $Rsense$, modelled as parasitic resistances. This results in a $G_{non-ideal}$ matrix, with each element $G_{ij}'$ incorporating the effect due to the non-idealities, obtained using circuit laws (Kirchoff's laws and Ohm's law) and linear algebraic operations~\cite{rxnn} (described in \figurename{~\ref{xbar}} (b)). Consequently, the net output current sensed at each column $j$ becomes $I_{j(non-ideal)} = \Sigma_{i=1}^{N}{G_{ij}' * V_i}$, which deviates from its ideal value. This manifests as accuracy degradation for DNNs mapped onto crossbars. So far, crossbar non-idealities have been projected in a negative light since they result in computational accuracy loss for DNNs. However, in this work, we show that greater the impact of non-idealities, more is the resilience of the mapped DNN towards adversarial attacks.
\section{Experiments and Results}
\subsection{Improving adversarial robustness of DNN using bit-error noise in hybrid memories}
In this section, we discuss how bit-error noise due to the erroneous 6T-SRAM cells at scaled supply voltages can be leveraged to improve the adversarial robustness of DNNs.  

\begin{figure}[t]
\centering
\includegraphics[width=0.3\textwidth]{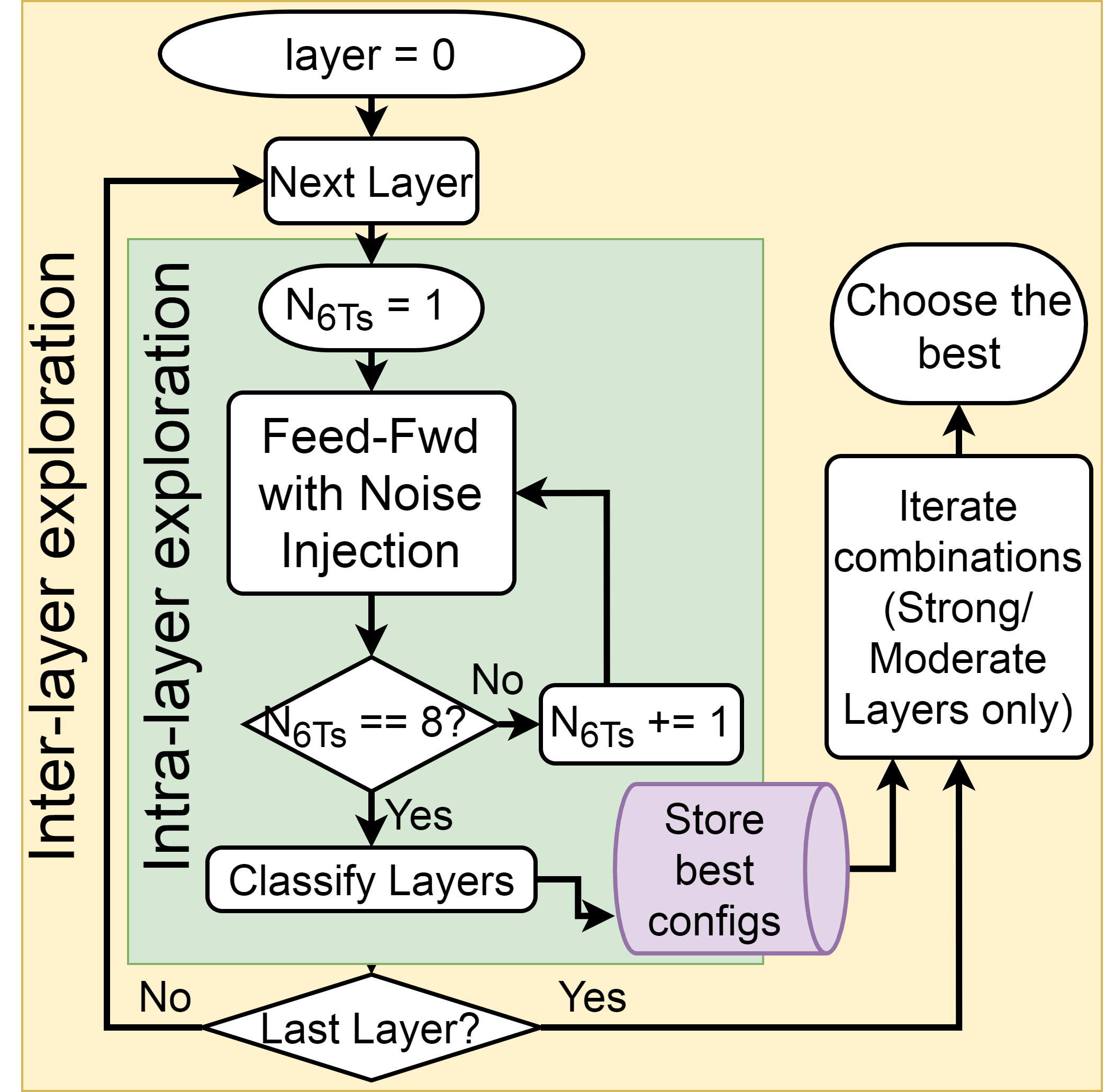}
\caption{Methodology for selecting the DNN layers suitable for bit-error injection and determining required $\mu$ values for the suitable layers. $\mu$ is obtained via choosing the right hybrid memory configurations shown in Fig. \ref{sn_vs_r_vs_vdd}}
\label{algo_rob}
\vspace{-4mm}
\end{figure}

In order to improve the adversarial robustness of the DNN using bit-error noise, we deliberately introduce controlled amounts of bit-error noise into hybrid memories of specific, strategically chosen DNN layers selected using the proposed methodology shown in Fig. \ref{algo_rob}. Note, the bit-error noise can be introduced either into the hybrid memories storing the activations of a layer or into the hybrid memories storing the parameters of the corresponding layer. We observe that introducing bit-error noise into the hybrid memories storing the layer activations offer higher adversarial accuracy improvements than introducing noise into the parameters. 

\textbf{Proposed methodology:} The goal of the proposed methodology shown in Fig. \ref{algo_rob} is to determine- i) the suitable layers for bit-error noise injection and, ii) the required amounts of bit-error noise in order to achieve significant adversarial accuracy improvements. The amount of bit-error noise is controlled using the hybrid memory configurations ($V_{DD}$ and $r$). First, the proposed methodology traverses through all the DNN layers. To determine the suitable amount of bit-error noise in each layer, the number of 6T-SRAM cells are incremented from 1 to 8 while keeping the $V_{DD}$ constant. Note, that the baseline model without any bit-error noise is trained using 8-bit activations and weights. Corresponding to each bit-error noise in each layer, we launch FGSM attack of a fixed strength (determined by $\epsilon$) on the modified DNN. Based on the adversarial accuracy values corresponding to each hybrid memory configuration, we select the configuration that yields the highest adversarial accuracy. Similarly, layers that yield more than 5\% higher adversarial accuracy than the baseline model with their corresponding best configurations are selected for the final step evaluation. 

In the final step evaluation, the shortlisted layers are combined in different ways and FGSM attack is launched on each combination separately. The combination that yields the best adversarial accuracy is selected and their corresponding best hybrid memory configurations are used. Note, throughout the entire methodology, we do not consider bit-error noise during the gradient calculation step for FGSM attack. 

\textbf{Layer specific hybrid memory configurations:} Table~\ref{layer_config vgg} and Table~\ref{layer_config resnet} show the hybrid memory configurations for the layers of VGG19 and ResNet18 networks. Based on the above methodology, the best hybrid memory configurations for the selected layers have been mentioned. Layers having H indicate that the activation memory of the corresponding DNN layer is comprised of homogeneous 8T or 6T SRAM memory. Additionally, P's in Table~\ref{layer_config vgg} denote pooling layers and S's in Table~\ref{layer_config resnet} denote shortcut layers of the residual network. Interestingly, we find that initial layers of both the networks, offer higher adversarial robustness than the deeper layers.

The last columns of both the tables depict the clean accuracy (CA) of the DNN with bit-error noise injected. Additionally, the deviation of the CA with respect to the CA of the baseline software model (without any bit-error noise) has been mentioned. A slightly lower value of CA is attributed to the regularization effect due to the bit-error noise introduced into the hybrid activation memories. Re-training the bit-error noise injected DNN with clean images can improve the CA of the network. 

\subsubsection{CIFAR-10 and CIFAR-100 results} 
For both CIFAR10 and CIFAR100 datasets, we launch FGSM attacks on both VGG19 and ResNet18 networks. During the FGSM attack, we do not include bit-error noise in the gradient calculation step. We introduce adversarial attacks of strengths ($\epsilon$) in an incremental fashion from 0.05 to 0.3. For both the datasets, introduction of bit-error noise into the selected layers shown in Table~\ref{layer_config vgg} and Table~\ref{layer_config resnet} lead to a decrease in the adversarial loss (AL) which means higher adversarial robustness. Moreover, among the two network architectures, VGG19 network shows an overall lower AL than ResNet18 networks. 
\begin{table*}[t]
\centering
\caption{Layer wise activation memory configurations with the corresponding 8T-6T values for the VGG19 network. P denotes the pooling layers. H denotes homogeneous activation memory hence no 8T-6T ratios have been mentioned}
\label{layer_config vgg}
\resizebox{\textwidth}{!}{%
\begin{tabular}{|c|c|c|c|c|c|c|c|c|c|c|c|c|c|c|c|c|c|c|c|c|c|c|c|}
\hline
Layer                         & 0 & 1   & \begin{tabular}[c]{@{}c@{}}2 \\ (P)\end{tabular} & 3 & 4   & \begin{tabular}[c]{@{}c@{}}5\\ (P)\end{tabular} & 6 & 7 & 8 & 9 & \begin{tabular}[c]{@{}c@{}}10\\ (P)\end{tabular} & 11 & 12 & 13 & 14 & \begin{tabular}[c]{@{}c@{}}15\\ (P)\end{tabular} & 16 & 17 & 18 & 19 & \begin{tabular}[c]{@{}c@{}}20\\ (P)\end{tabular} & $V_{DD}$ & \begin{tabular}[c]{@{}c@{}}Clean Accuracy/Deviation\end{tabular} \\ \hline
\multicolumn{1}{|l|}{CIFAR10} & H & 3/5 & 2/6                                              & H & 5/3 & H                                               & H & H & H & H & H                                               & H & H & H & H & H                                                & H & H & H & H & H                                                & 0.68V    & 88.78 / 2.61                                                       \\ \hline
CIFAR100                      & H & 3/5 & 2/6                                              & H & H   & 5/3                                             & H & H & H & H & H                                                & H & H & H & H & H                                                & H & H & H & H & H                                                & 0.68V    & 67.3 / 2.9                                                        \\ \hline
\end{tabular}%
}
\end{table*}
\begin{table*}[t]
\centering
\caption{Layer wise activation memory configurations with the corresponding 8T-6T values for the ResNet18 network. S denotes the shortcut layers. H denotes homogeneous activation memory hence no 8T-6T ratios have been mentioned}
\label{layer_config resnet}
\resizebox{\textwidth}{!}{%
\begin{tabular}{|c|c|c|c|c|c|c|c|c|c|c|c|c|c|c|c|c|c|c|c|c|c|c|c|c|c|c|}
\hline
Layer                         & 0 & 1   & \begin{tabular}[c]{@{}c@{}}2 \\ (S)\end{tabular} & 3 & 4   & \begin{tabular}[c]{@{}c@{}}5\\ (S)\end{tabular} & 6 & 7 & \begin{tabular}[c]{@{}c@{}}8\\ (S)\end{tabular} & 9 & 10 & \begin{tabular}[c]{@{}c@{}}11\\ (S)\end{tabular} & 12 & 13 & \begin{tabular}[c]{@{}c@{}}14\\ (S)\end{tabular} & 15 & 16 & \begin{tabular}[c]{@{}c@{}}17\\ (S)\end{tabular} & 18 & 19 & \begin{tabular}[c]{@{}c@{}}20\\ (S)\end{tabular} & 21 & 22 & \begin{tabular}[c]{@{}c@{}}23\\ (S)\end{tabular} & $V_{DD}$ & \begin{tabular}[c]{@{}c@{}}Clean Accuracy/Deviation\end{tabular} \\ \hline
\multicolumn{1}{|l|}{CIFAR10} & H & 4/4 & 5/3                                              & H & 6/2 & H                                               & H & H & H & H & H                                                & H & H & H & H & H                                                 & H & H & H & H & H  & H & H & H                                               & 0.68V    & 89.2 / 6.14                                                       \\ \hline
CIFAR100                      & 5/3 & H & 6/2                                              & 6/2 & H & H                                             & H & H & H & H & H                                                & H  & H  & H  & H  & H                                                & H  & H  & H  & H  & H & H & H & H                                               & 0.68V    & 69.4 / 7.1                                                        \\ \hline
\end{tabular}%
}
\end{table*}
\begin{figure}[t]
\centering
\includegraphics[width=0.5\textwidth]{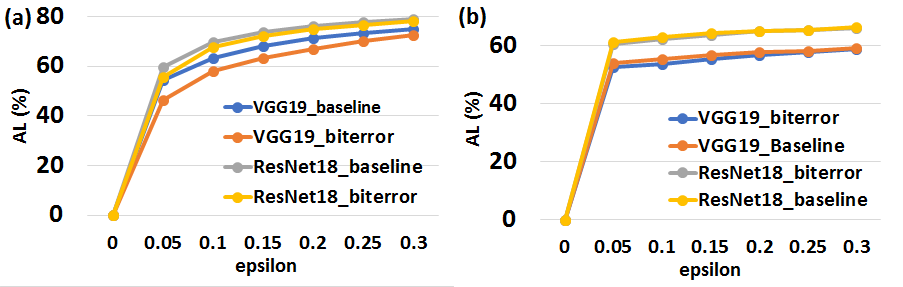}
\caption{Variation of AL with FGSM perturbation strength epsilon ($\epsilon$) for a) CIFAR10 and b) CIFAR100 with bit-error noise on both VGG19 and ResNet18 network architectures. Adversarial loss for \textit{Baseline} models is calculated without bit-error noise into the DNN network.}
\label{results}
\vspace{-4mm}
\end{figure}
\subsection{Harnessing robustness from non-idealities in crossbars}

\begin{figure*}[t]
    \centering
    \includegraphics[width=\linewidth]{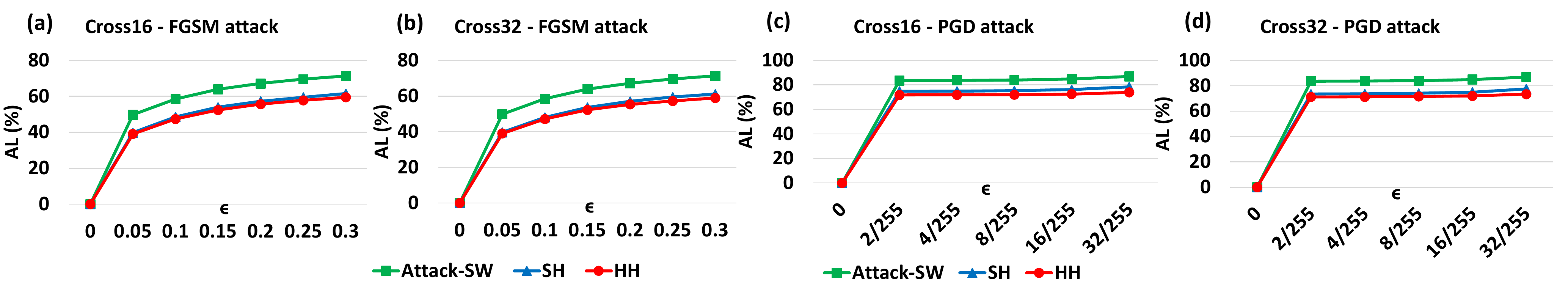}%
    \caption{(a)-(b) A plot between AL and $\epsilon$ for Attack-SW, SH and HH attacks (FGSM) on VGG8 with CIFAR-10 dataset for crossbar sizes 16x16 and 32x32; (c)-(d) A plot between AL and $\epsilon$ for Attack-SW, SH and HH attacks (PGD) on VGG8 with CIFAR-10 dataset for crossbar sizes 16x16 and 32x32}
    \label{normal-vgg8}
    \vspace{-5mm}
\end{figure*}

We conduct the crossbar robustness experiments on state-of-the-art VGG8 and VGG16 DNN architectures with benchmark datasets- CIFAR-10 \& CIFAR-100 respectively. After training the DNNs on software, we launch FGSM and PGD attacks by adding adversarial perturbations to the clean test inputs and record the adversarial accuracies in each case. We classify these attacks under Attack-SW. This forms our baseline software models.

We consider memristive crossbars with an ON/OFF ratio of 10 (\textit{i.e,} $R_{MIN} = 20 k\Omega$ and $R_{MAX} = 200 k\Omega$), having the resistive non-idealities as follows: $Rdriver = 1 k\Omega$, $Rwire\_row = 5 \Omega$, $Rwire\_col = 10 \Omega$ and $Rsense = 1 k\Omega$. Further, the device-level process variation has been modelled as a Gaussian variation in the conductances of the synaptic devices with $\sigma/\mu = 10\%$. In this work, we employ a \textit{PyTorch} based framework similar to \textit{RxNN} to map a DNN onto crossbar arrays (methodology explained in \figurename{~\ref{xbar}} (b)) to assess the impact of the inherent non-idealities on adversarial robustness of the mapped network. 

\textbf{Modes of adversarial attack: } We consider two modes of attack for the crossbar-mapped models of the DNNs - (a) \textit{Software-inputs-on-hardware (SH) mode} where, the adversarial perturbations for each attack are created using the software-based baseline's loss function and then added to the clean input that yields the adversarial input. The generated adversaries are then fed to the crossbar-mapped DNN. (b) \textit{Hardware-inputs-on-hardware (HH) mode} where, the adversarial inputs are generated for each attack using the loss from the crossbar-based hardware models. It is evident that HH perturbations will incorporate the effect of the intrinsic hardware non-idealities.

\subsubsection{CIFAR-10 results}
\label{sec:cifar10}

\begin{table}[t]
\centering
\caption{Table showing ALs (\%) in case of HH attack (PGD) on crossbar sizes of 16x16, 32x32 and 64x64 on a VGG8 network with CIFAR-10 dataset}
\label{cross}
\begin{tabular*}{\linewidth}{l@{\extracolsep{\fill}}@{}cccccc@{}}
\toprule

$\epsilon$ & 2/255 & 4/255 & 8/255 & 16/255 & 32/255 \\ \midrule
\textbf{Cross16} & 71.78 & 71.89 & 71.97 & 72.52 & 73.92 \\
\textbf{Cross32} & 71.13 & 71.22 & 71.55 & 71.97 & 73.36 \\
\textbf{Cross64} & 67.92 & 67.14 & 67.38 & 68.73 & 71.06\\ \bottomrule
\end{tabular*}
\vspace{-4mm}
\end{table}

In \figurename{~\ref{normal-vgg8}}, it can be observed that ALs in case of an adversarial attack on the DNNs mapped onto crossbars of sizes 16$\times$16 and 32$\times$32 are significantly lesser ($\sim10-15\%$) than those for the software baseline for different perturbation strengths ($\epsilon$). This is true for both FGSM and the stronger PGD attacks. In other words, the hardware-based non-idealities that come into play when DNNs are mapped onto crossbars provide robustness against adversarial inputs. Interestingly, we also find that larger crossbar sizes provide greater robustness against adversarial attacks (characterized by ALs for a given $\epsilon$) than the smaller ones. This is because larger crossbars involve greater number of parasitic components (non-idealities), thereby imparting more robustness. This can be seen in Table~\ref{cross} where, the 64$\times$64 crossbar shows least AL for a given $\epsilon$ implying better adversarial robustness. 

\textbf{Effect of $R_{MIN}$ on adversarial robustness: }The effective resistance of a crossbar structure is the parallel combination of resistances along its rows and columns. Earlier works such as~\cite{geniex} have shown that a smaller value of $R_{MIN}$ reduces the effective resistance of the crossbar and induces greater non-idealities. Thus, we expect crossbars with smaller $R_{MIN}$ values to impart better adversarial robustness to DNNs mapped onto them for a given ON/OFF ratio. We find that on decreasing $R_{MIN}$ to 10 $k\Omega$ (maintaining a constant $R_{MAX}/R_{MIN}$ ratio of 10), the ALs (for a PGD attack) in case of smaller $R_{MIN}$ are lower than the corresponding ALs for a larger $R_{MIN}$ as shown in \figurename{~\ref{compare}(a)}.

\subsubsection{CIFAR-100 results}
\label{sec:cifar100}

\begin{figure*}[t]
    \centering
    \includegraphics[width=\linewidth, page=1]{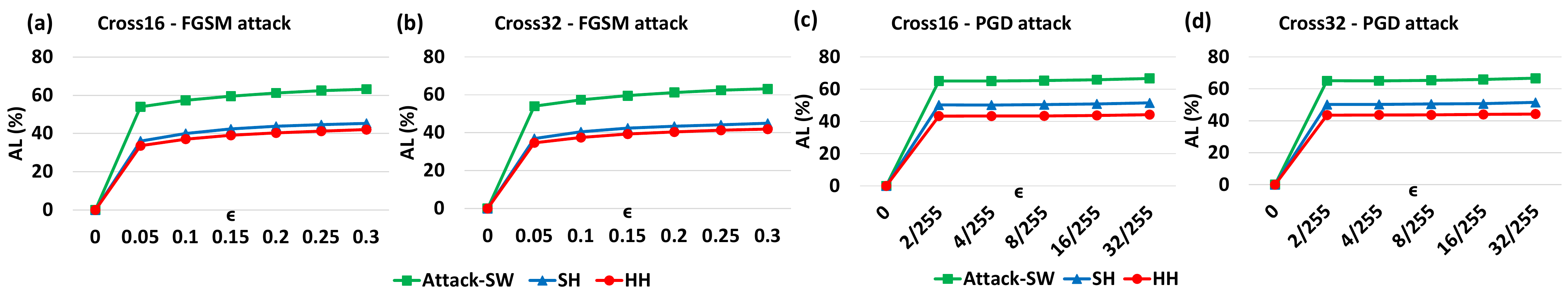}%
    \caption{(a)-(b) A plot between AL and $\epsilon$ for Attack-SW, SH and HH attacks (FGSM) VGG16 with CIFAR-100 dataset for crossbar sizes 16x16 and 32x32; (c)-(d) A plot between AL and $\epsilon$ for Attack-SW, SH and HH attacks (PGD) VGG16 with CIFAR-100 dataset for crossbar sizes 16x16 and 32x32}
    \label{vgg16}
\end{figure*}

\begin{figure*}[t]
    \centering
    \includegraphics[width=0.9\linewidth]{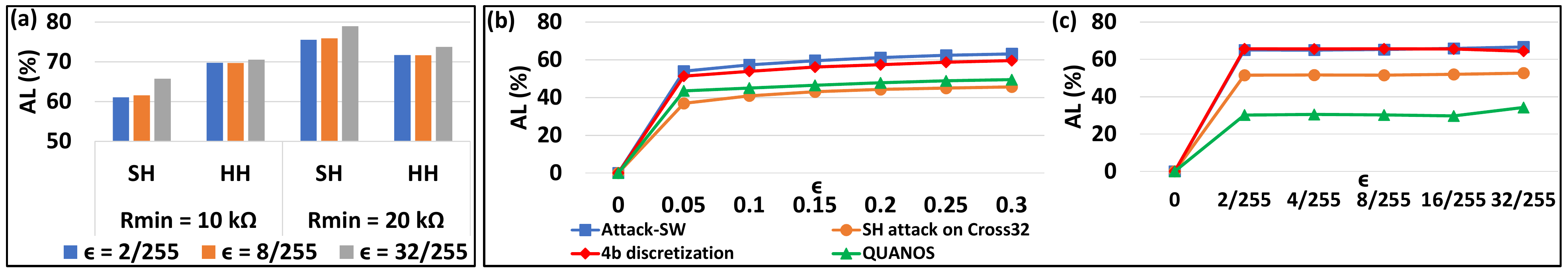}%
    \caption{(a) Bar-diagram showing ALs in case of SH and HH attacks (PGD) for a VGG8 network mapped on 32x32 crossbars using CIFAR-10 dataset for two different values of $R_{MIN}$ at constant $R_{MAX}/R_{MIN}$ ratio; (b)-(c) Comparison of our proposed method with other state-of-the-art adversarial defenses using VGG16 network and CIFAR-100 dataset during FGSM \& PGD attacks respectively}
    \label{compare}
\end{figure*}

The results shown in \figurename{~\ref{vgg16}} are similar to those for CIFAR-10 dataset. Crossbar-based non-idealities impart adversarial robustness to the mapped VGG16 network ($>10-20\%$) against both FGSM and PGD attacks. However, with a more complex CIFAR-100 dataset, we clearly observe that DNN shows greater adversarial robustness against PGD attack for HH attack \textit{w.r.t} SH attack ($>7\%$) than what is observed with CIFAR-10 dataset ($>4\%$).

\textbf{Comparison with Related works: }We compare the performance of non-ideality-driven adversarial robustness in crossbars against state-of-the-art software defenses described in~\cite{pixeld, quanos}. Note, \cite{pixeld, quanos} use efficiency driven transformations (that \textit{implicitly} translate to hardware benefits) such as, quantization to improve resilience. In contrast, our work utilizes \textit{explicit} hardware variations to improve robustness. We observe that for single-step FGSM SH mode attack on a VGG16 network mapped on 32x32 crossbars, adversarial robustness due to crossbar non-idealities outperforms all other techniques (\figurename{~\ref{compare}(b)}). For multi-step PGD attack, SH ranks second (\figurename{~\ref{compare}(c)}). With respect to 4-bit (4b) \textit{pixel discretization} of input data~\cite{pixeld}, non-idealities in crossbars impart $\sim15\%$ greater adversarial robustness in case of FGSM attack and $\sim12\%$ greater adversarial robustness in case of PGD attack. In case of FGSM attack, crossbar-based non-idealities impart $\sim4\%$ greater adversarial robustness than QUANOS~\cite{quanos}, while for PGD attack, QUANOS outperforms by $\sim18-22\%$.

\section{Conclusion}

This work presents an overview of how hardware implementation of DNNs guarantees better adversarial robustness against baseline software models. This is primarily due to the interference of intrinsic hardware noise on the creation of useful gradients responsible for unleashing adversarial attacks. We separately explore the effects of bit-error noise in hybrid 8T-6T CMOS memories as well as non-idealities pertaining to analog crossbar arrays on the robustness of DNNs and determine useful parameters on which the noise (or non-idealities) largely depend on. We finally validate our proposed ideas using CIFAR10 \& CIFAR100 datasets on VGG and ResNet architectures and present a comparison between the performance of hardware noise-driven robustness in hybrid 8T-6T SRAM memories and crossbars against state-of-the-art efficiency-driven quantization techniques to improve adversarial robustness. 
\vspace{-2mm}

\section*{Acknowledgement}
This work was supported in part by the National Science Foundation, the Amazon Research Award, and the Technology Innovation Institute, Abu Dhabi.
\vspace{-2mm}

\bibliographystyle{IEEEtran}
\bibliography{bib}

\end{document}